\documentclass[letterpaper, 10 pt, conference]{ieeeconf}

\IEEEoverridecommandlockouts
\usepackage{balance}
\usepackage{amssymb}
\usepackage{graphics}
\usepackage{graphicx}
\usepackage{amsmath}
\usepackage{floatrow}
\usepackage{url}
\usepackage{cite}
\usepackage{textcomp}
\usepackage{gensymb}
\usepackage[ruled,vlined]{algorithm2e}
\usepackage{multirow}
\usepackage{tablefootnote}
\usepackage{siunitx}
\usepackage{tabularray}
\usepackage{multicol}

\title{\LARGE \bf
Ablation Study of Multimodal Perception, Language Grounding, and Control for Human-Robot Interaction in an Object Detection and Grasping Task
}

\author{Tian Zi, Shen Guanting%
\thanks{The authors are with the Dalian University of Technology, Dalian, China. {\tt\footnotesize \{zi.tian.robotics, shen.guanting.99\}@gmail.com}. The first author is the corresponding author.}%
}

\begin{document}

\maketitle
\thispagestyle{empty}
\pagestyle{empty}

\begin{abstract}
This manuscript extends our previous multimodal human-robot interaction system by introducing a controlled ablation study of the three modules that most strongly influence end-to-end performance: the large language model used for action extraction, the perception system used for visual grounding, and the controller used for motion execution. The goal is not to redesign the full pipeline, but to isolate the contribution of each component under a common experimental protocol and then evaluate the best combinations end-to-end. We therefore compare three language models, five perception configurations, and three controllers, followed by a second-stage factorial study over the best candidates. The resulting analysis is intended to clarify which choices primarily affect execution time, which primarily affect success rate, and where the largest engineering gains are likely to come from in future revisions of the system.
\end{abstract}


\section{Introduction}\label{sec:introduction}

Human--robot interaction (HRI) has become a central research direction as robotic systems move from isolated automation tools toward collaborative agents that assist people in healthcare, education, industrial settings, and domestic environments~\cite{hu2011advanced,zhao2022human,he2017educational,kagami2006home}. In these scenarios, the robot is no longer expected to execute only preprogrammed motions; instead, it must interpret human goals, adapt to changing contexts, and respond in a manner that is both effective and intuitive. This shift places intention recognition~\cite{dominguez2025human} at the center of HRI design, because the quality of the interaction depends directly on how well the system can infer what the user wants and translate that intent into safe and useful action~\cite{huang2016anticipatory,lorentz2023pointing}.

The challenge becomes even more pronounced when the task involves physical collaboration at close range. In applications such as robotic surgery, errors in intent interpretation may have serious consequences for patient safety~\cite{saeidi2019autonomous}. Likewise, in human--robot collaborative transport and handover tasks, the robot must reason about force, motion, and timing while remaining responsive to human behavior~\cite{dominguez2025perception,dominguez2024exploring,dominguez2024force}. These lines of work show that successful HRI requires more than accurate perception alone: it requires a control strategy that can combine sensing, prediction, and action selection under uncertainty.

A related trend in HRI is the growing use of language and multimodal reasoning to move away from rigid command interfaces. Early efforts already explored multimodal and anticipatory interaction, such as language-and-sketch-based robot navigation~\cite{zu2024language}, interactive navigation with large language and vision--language models~\cite{zhang2024interactive}, and physically grounded vision--language reasoning for manipulation~\cite{gao2024physically}. At the same time, the broader HRI literature has emphasized that human intention is not always directly observable and may require inference from partial cues, context, and explicit dialogue~\cite{dominguez2025inference,dominguez2024anticipation}. These observations motivate systems that can interpret both speech and visual context rather than relying on a single modality.

Recent advances in large language models (LLMs) and vision--language models (VLMs) have made this direction substantially more practical. General-purpose models such as LLaMA and GPT-4, together with more recent instruction-tuned systems, provide strong natural language understanding and structured reasoning capabilities~\cite{touvron2023llama,achiam2023gpt}. In robotics, these models have been used to extract action-oriented meaning from user instructions and to support higher-level planning over multimodal scene information. However, the HRI community has also noted important pitfalls: language models can be overconfident, semantically brittle, or poorly grounded if their outputs are not checked against the physical world~\cite{atuhurra2024leveraging,dominguez2025inference}. As a result, the most effective systems are often hybrid architectures in which foundation models are paired with deterministic perception and control components.

Within that hybrid paradigm, open-vocabulary vision models have become especially appealing because they reduce the need for task-specific object detectors. Florence-2, for example, offers a unified representation for multiple vision tasks and can be prompted for open-vocabulary detection~\cite{xiao2024florence}. On the audio side, Whisper provides robust speech recognition from weak supervision at scale~\cite{radford2023robust}, while Audio Spectrogram Transformer (AST)-based models remain attractive for low-latency wake-word detection due to their compactness and specialization in short commands~\cite{gong2021ast}. On the control side, fuzzy logic remains a practical way to manage uncertainty and nonlinearity in robotic motion, especially when sensor noise and pose estimation errors are unavoidable~\cite{mendel2002fuzzy,liang2000interval,hellmann2001fuzzy,hagras2004type}. Taken together, these tools suggest a modular architecture in which perception, language understanding, and control can be combined without requiring a fully end-to-end learned robot policy.

Motivated by these developments, our previous work~\cite{shen2026approach} presented a multimodal HRI framework that coupled speech recognition, language understanding, open-vocabulary object perception, and fuzzy control to enable spoken object-manipulation commands for a Dobot Magician robotic arm. That study demonstrated the feasibility of the overall pipeline and achieved an end-to-end task success rate of 75\%, but it also revealed clear bottlenecks in both perception and action execution. In particular, the experiments showed that the overall performance was strongly affected by the specific choices made for the language model, the vision subsystem, and the controller, indicating that the system’s behavior was not determined solely by the high-level architecture but also by the capabilities of each module.

This observation motivates the present paper. Rather than introducing a new architecture from scratch, we take the previously established system as a baseline and ask a more focused question: which component choices most strongly influence performance, and under what conditions do they matter? To answer this, we conduct a systematic ablation study across three critical stages of the pipeline: the action-extraction LLM, the perception subsystem, and the motion controller. By comparing alternative models and control strategies under controlled experimental conditions, this paper aims to isolate the contribution of each stage, quantify the latency--accuracy trade-offs, and identify the most promising configurations for future work. We believe that this comparative perspective is essential for turning a functional prototype into a more reliable and scientifically grounded HRI system.

The main contributions of this paper are threefold. First, we provide a controlled benchmark of multiple LLMs for action extraction in spoken HRI commands. Second, we compare several perception pipelines for object localization and grasp-point estimation under identical task conditions. Third, we evaluate alternative controllers to determine how different motion-generation strategies affect execution time and task success. The remainder of the paper is organized as follows: Section~\ref{sec:baseline} revisits the baseline system and defines the experimental setup; Section~\ref{sec:design} describes the ablation protocol; Section~\ref{sec:results} reports the quantitative results; Section~\ref{sec:discussion} discusses the findings and their implications; Section~\ref{sec:limitations} presents the limitations; and Section~\ref{sec:conclusion} concludes the paper and outlines future directions.

\section{Baseline System Recap}
\label{sec:baseline}

This paper builds directly on the multimodal manipulation framework introduced in~\cite{shen2026approach}. Because the scientific objective of the present study is to isolate the effect of alternative language models, perception pipelines, and controllers, this section only restates the elements that remain fixed across the ablation study. Full implementation details of the baseline system---including the internal configuration of the speech, vision, and control modules---are reported in the previous article and are not repeated here unless they are required to understand the experimental comparisons.

The baseline system is a closed-loop human--robot interaction pipeline in which a spoken instruction is converted into a structured action, grounded in the scene, and executed by a Dobot Magician arm. Across all experiments, the same physical platform, the same camera and audio hardware, the same workspace geometry, the same object vocabulary, the same calibration procedure, and the same trial protocol are preserved. Only the module under study is changed in each experimental round; all other stages remain identical to the baseline configuration described in~\cite{shen2026approach}.

\subsection{Hardware Platform and Fixed Infrastructure}

The experimental platform is centered on a Dobot Magician robotic arm equipped with the same suction-based end-effector used in the first article. Visual sensing is provided by an Intel RealSense D435i RGB-D camera mounted at a fixed viewpoint relative to the tabletop, while spoken commands are captured using Samsung Buds2 wireless earbuds. All computation is performed locally on the same laptop platform, an Intel Core i9-14900HX machine with an NVIDIA RTX 4070 GPU, so that inference latency, model loading behavior, and control timing are measured under the same hardware constraints in every trial.

To preserve comparability across all runs, the study keeps the camera intrinsics, camera placement, camera-to-table distance, robot base pose, workspace boundaries, and scene layout unchanged. The same tabletop environment is used throughout the ablation study, with the same fruit-based object vocabulary and the same arrangement protocol inherited from the baseline paper. In addition, the same local software stack is retained: the command queue, logging utilities, graphical user interface, and safety logic operate identically across all conditions, and only the module being evaluated is replaced from one condition to another.

\begin{figure}[t]
	\centering
	\includegraphics[width=0.98\textwidth]{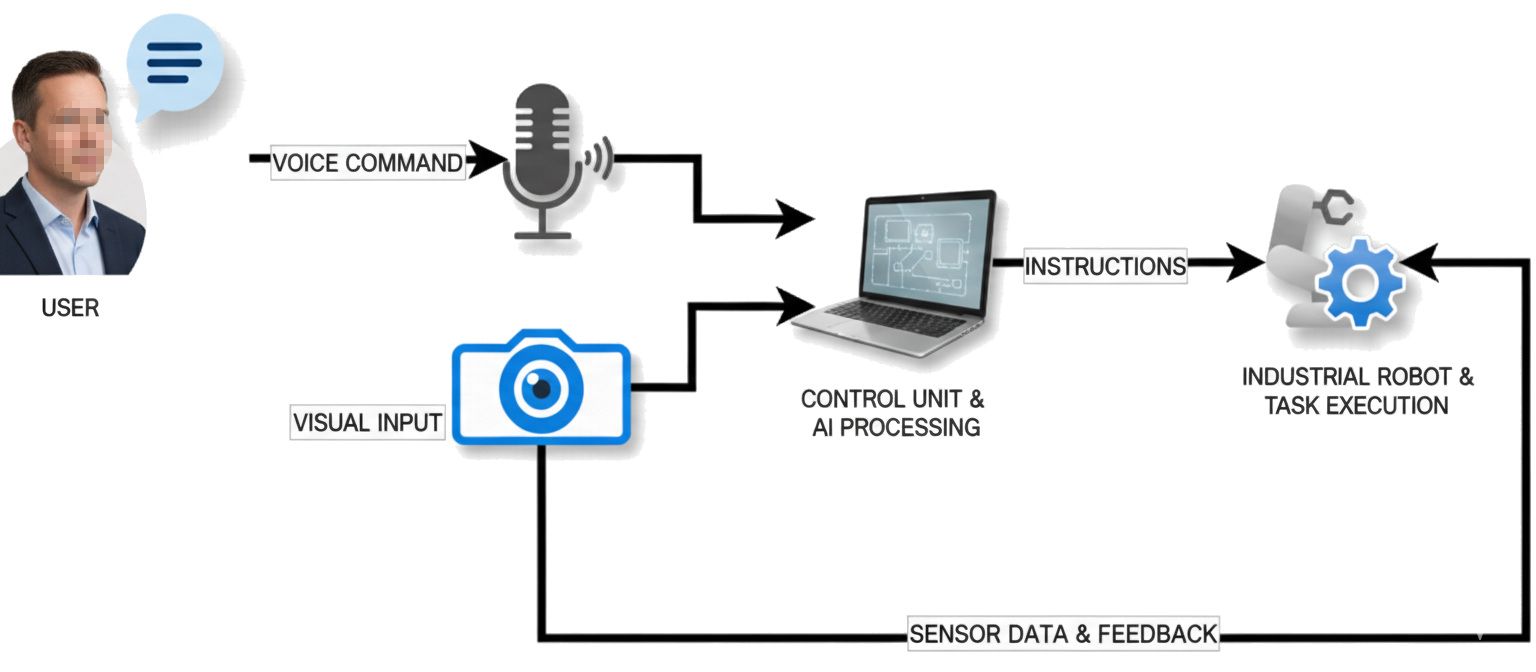}
	\caption{{\bf Information flow of the complete pipeline.} Hardware elements and data flow of the robotic manipulation system. Figure extracted from~\cite{shen2026approach}.}
	\label{fig:Block-diagram}
\end{figure}

\begin{table}[t]
    \centering
    \caption{Baseline platform elements held constant during the ablation study.}
    \label{tab:baseline-elements}
    \begin{tabular}{p{0.26\linewidth} p{0.64\linewidth}}
        \hline
        \textbf{Element} & \textbf{Kept fixed across all experiments} \\
        \hline
        Robot arm & Dobot Magician with the same suction-based end-effector, robot base pose, and motion workspace used in the baseline article. \\
        Vision sensor & Intel RealSense D435i mounted in the same position and orientation, with unchanged intrinsics, extrinsics, and capture resolution. \\
        Audio input & Samsung Buds2 wireless earbuds used as the same speech input device, with the same wake-word and recording protocol. \\
        Compute unit & The same local laptop platform (Intel Core i9-14900HX + NVIDIA RTX 4070) and the same software environment for model execution. \\
        Workspace & The same tabletop scene, camera-to-table distance, lighting conditions, and fruit-object arrangement protocol used in the first article. \\
        Software stack & The same command queue, GUI, logging logic, safety checks, and trial control code; only the module under test changes across conditions. \\
        \hline
    \end{tabular}
\end{table}

\subsection{Fixed Interaction Pipeline}

The interaction loop remains the same in every round of the ablation study. A participant issues a spoken command, the audio front-end detects the wake word, and the resulting utterance is transcribed into text. The transcription is then passed to the language module, which maps the command to a structured action representation. In parallel, the perception module identifies the referenced object(s) in the scene, and the controller computes the motion required for the robot to move toward the target. The robot then executes the commanded manipulation while the system logs time stamps, intermediate outputs, and the final success state.

In the first experimental round, only the language model used in the action-extraction stage is changed, while the speech pipeline, perception pipeline, and controller are kept fixed. In the second round, only the perception system is replaced, and the speech pipeline, language model, and controller remain unchanged. In the third round, only the controller is varied. This structure ensures that each reported difference can be interpreted as the effect of one specific design choice rather than as the result of multiple interacting changes. The final combination round then tests the best-performing candidates from the earlier rounds under identical conditions.

\subsection{Task Definition and Command Space}

The task studied in this paper is spoken-command object manipulation in a controlled tabletop environment. The user issues a natural-language instruction referring to one of the objects in the scene, and the robot is expected to identify the correct target and execute the corresponding motion. The command space is intentionally kept the same as in the baseline article so that the new experiments are directly comparable: commands are short imperative utterances describing object identity and, when necessary, spatial relations such as left of, right of, next to, or in front of another object. The object vocabulary remains limited to the same fruit-based set used in the original system, which keeps the language-to-action mapping consistent across all experimental rounds.

A trial is considered successful only if the system completes the requested interaction end-to-end: the spoken command must be correctly processed, the intended object must be grounded in the scene, the robot must move to the correct target, and the commanded manipulation must be completed without a control failure or an incorrect object selection. Partial successes are still recorded at the module level, but the main metric reported later in the paper is the binary end-to-end outcome, as in the baseline study. Timing is measured from wake-word detection to the completion of the physical action, which allows direct comparison of latency across the different LLM, perception, and controller options.

\begin{table}[t]
    \centering
    \caption{Representative task and command-space characteristics used in the ablation study.}
    \label{tab:task-space}
    \begin{tabular}{p{0.28\linewidth} p{0.62\linewidth}}
        \hline
        \textbf{Item} & \textbf{Description used in the paper} \\
        \hline
        Command type & Short spoken imperatives requesting object pickup, placement, or relative repositioning in the tabletop scene. \\
        Objects & The same fruit-based object vocabulary from the baseline article, reused unchanged across all trials. \\
        Scene type & A controlled tabletop setup with fixed camera viewpoint, fixed calibration, and the same arrangement protocol in every run. \\
        Success criterion & Binary end-to-end success: correct command interpretation, correct object grounding, and successful physical execution of the requested action. \\
        Failure modes & Missed wake word, transcription error, malformed or incorrect action extraction, wrong object grounding, and control or execution failure. \\
        Logging & Stage-wise time stamps and success labels are recorded for the LLM, perception, controller, and final task outcome. \\
        \hline
    \end{tabular}
\end{table}

For readability and interpretability, the new ablation study retains the same evaluation philosophy as the baseline paper: each module is assessed both in isolation and in its contribution to the complete pipeline. The reader may therefore use the previous article~\cite{shen2026approach} for the fine-grained implementation details of AST-based wake-word detection, Whisper transcription, Florence-2 perception, and the interval Type-2 fuzzy controller, while the remainder of this paper focuses on how alternative design choices affect speed and success rate.

\section{Experimental Design}
\label{sec:design}

The experimental study is organized as a controlled ablation analysis of the end-to-end object-detection and grasping pipeline introduced in the baseline system~\cite{shen2026approach}. Rather than modifying the full architecture at once, we vary one component family at a time while keeping all remaining modules, hardware settings, calibration parameters, and user interaction rules fixed. This design makes it possible to attribute performance differences to a specific language model, perception system, or controller instead of to uncontrolled changes in the overall pipeline.

The study is divided into four rounds. In the first round, we compare three large language models (LLMs) used for action extraction while all other stages remain unchanged. In the second round, we compare five perception configurations while keeping the LLM and controller fixed. In the third round, we compare three controllers under the same upstream perception output. Finally, in the fourth round, we evaluate the best two candidates from each of the previous rounds in a full factorial combination study. In total, the protocol contains 380 independent trials: 60 for the LLM ablation, 100 for the perception ablation, 60 for the controller ablation, and 160 for the combined best-contenders study.

\subsection{Research Questions and Hypotheses}

The ablation study is designed to answer three main questions. First, which LLM yields the most reliable action extraction from spoken commands while keeping inference time low? Second, which perception configuration best balances localization accuracy and latency in the tabletop manipulation task? Third, which controller provides the best trade-off between smooth motion, robustness to uncertainty, and execution speed?

Before running the experiments, we expected the following trends. For the language models, stronger instruction-following ability was expected to reduce malformed action outputs and improve end-to-end success, although differences in latency were expected to remain modest. For the perception systems, methods with better spatial grounding or depth-aware post-processing were expected to improve grasp accuracy, but at the cost of additional runtime. For the controllers, the interval Type-2 fuzzy logic controller was expected to provide the most stable behavior under perception noise, whereas the PID baseline was expected to be fast but more sensitive to tuning and estimation error. The fourth round was intended to test whether the strongest single-module choices also remain best when combined.

\subsection{Experimental Matrix and Run Counts}

Table~\ref{tab:experimental-matrix} summarizes the overall structure of the study. Each candidate in a given round is evaluated using 20 trials under identical task conditions. This fixed number of repetitions was selected to provide a stable estimate of both success rate and time while keeping the total experimental burden manageable. The combined study then uses the shortlisted top two candidates from each module family, producing $2 \times 2 \times 2 = 8$ configurations and therefore 160 additional trials.

\begin{table*}[t]
    \centering
    \caption{Planned experimental matrix for the ablation study.}
    \label{tab:experimental-matrix}
    \begin{tabular}{l l c c c}
        \hline
        \textbf{Round} & \textbf{Variable factor} & \textbf{Candidates} & \textbf{Runs per candidate} & \textbf{Total runs} \\
        \hline
        1 & LLM for action extraction & 3 & 20 & 60 \\
        2 & Perception system & 5 & 20 & 100 \\
        3 & Controller & 3 & 20 & 60 \\
        4 & Top-$2 \times$ top-$2 \times$ top-$2$ combinations & 8 combinations & 20 & 160 \\
        \hline
        \multicolumn{4}{r}{\textbf{Total}} & \textbf{380} \\
        \hline
    \end{tabular}
\end{table*}

\subsection{Common Trial Protocol and Randomization}

All experiments follow the same interaction protocol. A participant issues a spoken command after the wake-word trigger has been detected, the speech stream is transcribed, the text is converted into a structured action representation, the relevant object is localized, and the robot executes the commanded motion. The tabletop scene, camera viewpoint, lighting conditions, calibration parameters, object set, and logging pipeline are kept constant across all rounds so that the only source of variation is the module under test.

To reduce ordering effects, the order of trials within each round is randomized across candidates. When repeated measurements are taken for the same candidate, the commands are distributed as evenly as possible so that no system benefits from a fixed command sequence or from a particular object arrangement. Between trials, the workspace is reset to the same nominal configuration. This ensures that the comparisons reflect model and controller behavior rather than accumulated drift in the environment.

\subsection{Evaluation Metrics}

The same set of metrics is reported in every round. Let $T_{AE}$ denote the time required by the LLM to extract the action sequence from the transcribed command, $T_{OD}$ the time required by the perception module to localize the target object(s), and $T_{RA}$ the time required by the controller and robot to complete the requested motion. The corresponding success indicators are denoted by $A_{AE}$, $A_{OD}$, and $A_{RA}$, respectively, where each module contributes a binary correctness value for the trial.

For an end-to-end task, the total completion time is measured from wake-word detection until the robot finishes the commanded action:
\begin{equation}
    T_{total} = T_{AE} + T_{OD} + T_{RA} + C,
\end{equation}
where $C$ captures constant overheads such as system synchronization, command dispatching, and logging. The end-to-end success metric $A_{total}$ is also binary: a trial is counted as successful only when the command is interpreted correctly, the intended object is grounded correctly, and the physical action is completed without failure. The primary reporting format in this paper is therefore mean success rate and mean completion time, together with their variability across the 20 runs per condition.

\begin{table}[t]
    \centering
    \caption{Metrics reported in every experimental round.}
    \label{tab:metrics}
    \begin{tabular}{p{0.22\linewidth} p{0.66\linewidth}}
        \hline
        \textbf{Metric} & \textbf{Meaning in the present study} \\
        \hline
        $T_{AE}$ & Time required by the language model to convert text into actions. \\
        $T_{OD}$ & Time required by the perception system to localize the target object(s). \\
        $T_{RA}$ & Time required by the controller and robot to execute the action. \\
        $A_{AE}$ & Correctness of the extracted action sequence. \\
        $A_{OD}$ & Correctness of the detected object(s) and grasp/target point. \\
        $A_{RA}$ & Correctness of the final physical execution. \\
        $A_{total}$ & End-to-end task success rate. \\
        $T_{total}$ & End-to-end completion time. \\
        \hline
    \end{tabular}
\end{table}

\subsection{Round 1: LLM Ablation Study}

In the first round, we isolate the action-extraction stage and compare the three LLMs used in the system. The speech transcription, perception, and controller modules are kept fixed so that any difference in performance can be attributed to the language model itself. Each LLM receives the same transcribed command and the same prompt template, which constrains the output to a structured action description that can be parsed directly by the robot controller. This setup allows us to measure both semantic correctness and inference time under identical task conditions.

The language-model candidates are the same ones used in the implementation described in the results section: Llama~3.1 8B, Ministral~3 8B, and Qwen~3.5 9B. For each model, 20 trials are executed with the same command set and the same parser. A candidate is considered successful in a trial only if the model returns a valid, executable action structure that matches the intended manipulation. Invalid formats, missing arguments, or semantically incorrect actions are counted as failures.

\subsection{Round 2: Perception Ablation Study}

The second round evaluates the visual grounding stage. In this round, the language model and the controller remain fixed, and only the perception subsystem is replaced. We compare five configurations: Florence-2, Florence-2 with depth-based centroid post-processing, fine-tuned YOLOv10, fine-tuned YOLOv10 with depth-based centroid post-processing, and Grounding DINO + SAM. Each configuration is tested on the same tabletop scene and the same object vocabulary so that the comparison reflects differences in localization quality and inference cost rather than differences in scene complexity.

For configurations that produce bounding boxes, the object position used by the controller is derived from the predicted box center or from the depth-refined centroid, depending on the method. For the segmentation-based configuration, the target point is obtained from the detected mask and then converted into the coordinate representation required by the controller. The evaluation records whether the correct object is selected, whether the target location is sufficiently accurate for successful grasping, and how long the perception step takes to complete. This round is therefore the main indicator of how vision quality propagates into downstream manipulation success.

\subsection{Round 3: Controller Ablation Study}

The third round compares the motion controller while keeping the upstream perception output fixed. This ensures that all controllers receive the same estimated target positions and that observed differences are due to control behavior rather than perception noise. The three controllers under test are Type-1 fuzzy logic control (Type-1 FLC), interval Type-2 fuzzy logic control (IT2FLS), and PID control.

Each controller is evaluated under the same motion constraints, safety limits, and stopping criteria. In particular, the robot is required to remain within the workspace boundaries, avoid excessive overshoot, and stop once the target is reached within the predefined tolerance region. The measured quantities are the time required to complete the motion and the binary success of the physical execution. This round is intended to reveal whether the extra uncertainty modeling of IT2FLS provides a measurable advantage over the simpler alternatives.

\subsection{Round 4: Combined Best-Contenders Study}

The fourth round combines the strongest candidates from the previous rounds and evaluates them jointly. The purpose of this stage is to determine whether the best individual module choices remain best when deployed together and to identify possible interaction effects between language understanding, visual grounding, and motion control. The shortlist is defined by ranking the candidates from each single-module study according to mean end-to-end success rate, using mean execution time as the primary tie-breaker and stability across trials as the secondary tie-breaker. The top two candidates from each family are then combined exhaustively, yielding eight configurations.

This factorial stage is important because module-wise superiority does not necessarily imply system-wise superiority. A candidate that performs best in isolation may interact poorly with the other modules, while a slightly weaker candidate may produce a better overall pipeline when combined with a more compatible perception or control strategy. By testing all pairwise combinations of the shortlisted candidates, the study captures these interactions while keeping the number of experiments manageable.

\begin{table}[t]
    \centering
    \caption{Selection rule for the second-stage factorial study.}
    \label{tab:selection-rule}
    \begin{tabular}{p{0.28\linewidth} p{0.62\linewidth}}
        \hline
        \textbf{Criterion} & \textbf{How it is applied} \\
        \hline
        Primary ranking & Highest mean success rate in the single-module study. \\
        Tie-breaker 1 & Lower mean execution time. \\
        Tie-breaker 2 & Lower run-to-run variance and fewer malformed outputs or control failures. \\
        Final shortlist & Top two candidates from each module family. \\
        \hline
    \end{tabular}
\end{table}

\subsection{Statistical Analysis and Reporting Protocol}

For each candidate, the 20-run sample is summarized by the sample mean, standard deviation, and observed range of the measured time and success values. Success rates are reported as percentages, and the end-to-end success metric is binary at the trial level. The primary comparison in each round is therefore descriptive: candidate systems are ordered by mean success rate and, when necessary, by mean time. This reporting convention keeps the focus on the practical behavior of each module family while remaining consistent with the experimental design.

When the full factorial study is analyzed, the same reporting protocol is used so that the combined configurations can be compared directly against the single-module baselines. The resulting tables and plots are intended to support a clear interpretation of which module choices matter most, where the largest time cost is introduced, and how strongly the different components interact when assembled into a complete robot manipulation pipeline.

\section{Results}
\label{sec:results}

This section reports the outcomes of the four experimental rounds in the same order used in the design protocol. For each candidate, we summarize the mean completion time and mean end-to-end success rate over 20 trials. The main objective is not only to identify the best-performing module in each family, but also to determine whether the single-module winners remain strong once they are combined into a complete pipeline.

\subsection{Round 1 Results: LLM Comparison}

Table~\ref{tab:llm-results} compares the three language models used for action extraction. All three candidates operated in a narrow latency range, with mean extraction times between 3.21~s and 3.50~s. This indicates that the language stage is not the dominant source of delay in the pipeline. The more important difference lies in success rate: Llama~3.1~8B reached 85\%, while Ministral~3~8B and Qwen~3.5~9B both achieved 90\%.

Among the two top performers, Ministral~3~8B was slightly faster than Qwen~3.5~9B by 0.29~s on average, so it provides the best speed--success trade-off in this round. Llama~3.1~8B remains competitive, but its lower success rate suggests that its structured action extraction is less reliable under the command format used in this study. Based on these results, Ministral~3~8B and Qwen~3.5~9B are retained for the combined factorial study.

\begin{table}[t]
    \centering
    \caption{Round 1 results template: language-model comparison.}
    \label{tab:llm-results}
    \begin{tabular}{lcc}
        \hline
        \textbf{LLM} & \textbf{Mean time (s)} & \textbf{Mean success (\%)} \\
        \hline
        Llama 3.1 8B & 3.46 & 85.0 \\
        Ministral 3 8B & 3.21 & 90.0 \\
        Qwen 3.5 9B & 3.50 & 90.0 \\
        \hline
    \end{tabular}
\end{table}

\subsection{Round 2 Results: Perception Comparison}

The perception study shows the strongest variation across candidates, both in latency and in success. As shown in Table~\ref{tab:vision-results}, Florence-2 achieved 80\% success with a mean time of 9.15~s, while adding depth-based centroid refinement improved the success rate to 85\% at the cost of additional latency. The fine-tuned YOLOv10 variants performed substantially better: the plain detector reached 90\% success in 5.69~s, and the version with depth centroid post-processing improved to 95\% success in 6.78~s. Grounding DINO + SAM also reached 95\% success, but with the highest runtime in the round at 11.91~s.

These results indicate that perception quality has a direct effect on downstream manipulation success. In particular, centroid refinement consistently improves the detector outputs, and the fine-tuned YOLOv10 pipeline offers a favorable balance between speed and accuracy. Grounding DINO + SAM matches the best success rate, but its additional runtime makes it less attractive when latency is considered. For this reason, YOLOv10 fine-tuned + depth centroid and Grounding DINO + SAM are selected for the factorial study.

\begin{table*}[t]
    \centering
    \caption{Round 2 results template: perception-system comparison.}
    \label{tab:vision-results}
    \begin{tabular}{lcc}
        \hline
        \textbf{Perception system} & \textbf{Mean time (s)} & \textbf{Mean success (\%)} \\
        \hline
        Florence-2 & 9.15 & 80.0 \\
        Florence-2 + depth centroid & 10.29 & 85.0 \\
        YOLOv10 fine-tuned & 5.69 & 90.0 \\
        YOLOv10 fine-tuned + depth centroid & 6.78 & 95.0 \\
        Grounding DINO + SAM & 11.91 & 95.0 \\
        \hline
    \end{tabular}
\end{table*}

\subsection{Round 3 Results: Controller Comparison}

The controller comparison in Table~\ref{tab:controller-results} shows a smaller spread in execution time, but a clear difference in robustness. Type-1 FLC and PID completed the motion in almost the same average time, 9.87~s and 9.85~s, respectively, while IT2FLS required 10.14~s. Despite this modest latency increase, IT2FLS achieved the highest success rate at 80\%, compared with 75\% for Type-1 FLC and 70\% for PID.

This pattern suggests that the benefit of interval Type-2 fuzzy control is not speed, but improved robustness under uncertainty. The extra modeling flexibility appears to reduce execution failures, which is consistent with the role of fuzzy control in handling imprecise perception and motion deviations. Although Type-1 FLC remains a strong baseline because of its slightly lower latency, IT2FLS is the preferred controller for the combined study because it provides the best overall success rate. PID is retained only as a lower-performing reference.

\begin{table}[t]
    \centering
    \caption{Round 3 results template: controller comparison.}
    \label{tab:controller-results}
    \begin{tabular}{lcc}
        \hline
        \textbf{Controller} & \textbf{Mean time (s)} & \textbf{Mean success (\%)} \\
        \hline
        Type-1 FLC & 9.87 & 75.0 \\
        IT2FLS & 10.14 & 80.0 \\
        PID & 9.85 & 70.0 \\
        \hline
    \end{tabular}
\end{table}

\subsection{Round 4 Results: Combined Best-Contenders Study}

The fourth round evaluates all combinations formed from the top two candidates of the previous three rounds, yielding eight end-to-end configurations. Table~\ref{tab:combined-results} reports the corresponding success rates. The main observation is that the best single-module choices do not automatically produce the best complete system. In several cases, changing only the controller or the perception module alters the final outcome by a noticeable margin, which confirms that the modules interact nontrivially once they are assembled into a full pipeline.

Among all eight combinations, Qwen~3.5~9B + Grounding DINO + SAM + IT2FLS achieves the highest success rate at 90\%. The same perception and controller pair with Ministral~3~8B attains 80\% to 85\% success depending on the controller, while the YOLOv10-based combinations remain in the 80\% to 85\% range. This result shows that the final system performance is not determined by a single dominant component; instead, the language model, perception module, and controller must be matched carefully.

From a selection standpoint, the combined study identifies Qwen~3.5~9B, Grounding DINO + SAM, and IT2FLS as the strongest overall configuration in terms of success rate. At the same time, the fact that some alternative combinations remain close in performance indicates that there is still room to trade off latency, robustness, and implementation complexity depending on the target application.

\begin{table*}[t]
    \centering
    \caption{Round 4 results template: combined best-contenders study.}
    \label{tab:combined-results}
    \begin{tabular}{cccc}
        \hline
        \textbf{LLM} & \textbf{Perception} & \textbf{Controller} & \textbf{Mean success} \\
        \hline
        Ministral 3 8B & YOLOv10 ft + d & Type-1 FLC & 80.0 \\
        Ministral 3 8B & YOLOv10 ft + d & IT2FLS & 80.0 \\
        Ministral 3 8B & Grounded-SAM & Type-1 FLC & 85.0 \\
        Ministral 3 8B & Grounded-SAM & IT2FLS & 80.0 \\
        Qwen 3.5 9B & YOLOv10 ft + d & Type-1 FLC & 80.0 \\
        Qwen 3.5 9B & YOLOv10 ft + d & IT2FLS & 85.0 \\
        Qwen 3.5 9B & Grounded-SAM & Type-1 FLC & 85.0 \\
        Qwen 3.5 9B & Grounded-SAM & IT2FLS & 90.0 \\
        \hline
    \end{tabular}
\end{table*}

\section{Discussion}
\label{sec:discussion}

The ablation study provides a clearer picture of where performance gains originate in the proposed human--robot interaction pipeline. Rather than improving uniformly when every module is changed, the system exhibits different sensitivities depending on whether the objective is speed, success rate, or overall robustness. The main empirical lesson is that the perception module has the strongest influence on end-to-end success, the controller has the most visible effect on execution stability, and the language model mainly affects semantic reliability with only modest latency differences. These trends are consistent with the stage-level measurements in Tables~\ref{tab:llm-results}--\ref{tab:combined-results} and help explain why the final system should not be selected by considering any single component in isolation.

\subsection{What Improved the System Most}

Among the three module families, perception produced the largest change in performance. Table~\ref{tab:vision-results} shows that success varied from 80\% to 95\% across the five perception configurations, and the corresponding runtime ranged from 5.69~s to 11.91~s. This spread is substantially larger than the variations observed for the language models and controllers. In practical terms, the quality of visual grounding determines whether the robot approaches the correct target in the first place, which then affects both grasp success and the total number of corrective motions required. The perception stage therefore acts as the principal gatekeeper for downstream performance.

The controller family had a smaller but still important effect. As reported in Table~\ref{tab:controller-results}, the difference between the fastest and slowest controllers was only a few tenths of a second, yet the success rate changed from 70\% for PID to 80\% for IT2FLS. This indicates that controller choice is less about saving time and more about absorbing uncertainty introduced by perception noise and small pose errors. In other words, the controller does not dominate the total runtime, but it strongly influences whether a detected object can actually be grasped reliably.

The language model stage contributed the least to total latency, with all three candidates clustered near 3.2--3.5~s in Table~\ref{tab:llm-results}. Even so, the success rate still changed from 85\% for Llama~3.1~8B to 90\% for both Ministral~3~8B and Qwen~3.5~9B. This suggests that the language model matters primarily through output validity and command structure rather than speed. For the command style used in this study, the important question is not which model generates text most quickly, but which one most consistently produces a parseable and task-consistent action description.

Taken together, these results imply a clear practical prioritization. If the goal is to improve robustness, perception should be addressed first, followed by the controller. If the goal is to reduce latency, the main opportunities lie in the visual pipeline rather than in the language model. The original 75\% end-to-end success rate reported in the baseline system is therefore not limited by a single weak component, but by the interaction of a relatively expensive perception stage with a control stage that must compensate for imperfect visual input.

\subsection{Failure Modes and Error Taxonomy}

The trial outcomes can be interpreted through a small set of recurring failure modes. The first and most consequential failure is wrong object grounding, where the system selects an incorrect instance or misses the intended target entirely. This error typically propagates through the full pipeline: once the wrong object is selected, the robot may move correctly but still fail the task. The fact that the perception configurations show the largest spread in success rate supports this interpretation.

A second failure mode is incomplete or malformed action extraction. In these cases, the language model produces a command that is syntactically valid but semantically ambiguous, partially specified, or inconsistent with the available controller methods. Such failures are especially damaging because they can prevent execution before the robot even begins to move. The comparatively narrow timing range in Table~\ref{tab:llm-results}, combined with the 5\% success gap between Llama~3.1~8B and the two stronger models, suggests that the language stage is mostly a correctness problem rather than a latency problem.

A third class of failure is motion-related. These errors include overshoot, oscillatory micro-corrections, and missed grasps caused by small residual pose errors. The controller results in Table~\ref{tab:controller-results} indicate that IT2FLS reduces these failures relative to Type-1 FLC and PID, which is consistent with the role of interval Type-2 modeling in absorbing uncertainty. In the combined study, the highest-success configuration pairs Grounding DINO + SAM with IT2FLS, which suggests that the most precise perception backend benefits most from the more robust controller.

\subsection{Speed-Accuracy Trade-offs}

The results also reveal a clear trade-off between speed and accuracy, especially in the perception subsystem. Florence-2 is slower than the fine-tuned YOLOv10 baseline and less accurate than the best configurations, while Grounding DINO + SAM matches the highest success rate in Round~2 but requires the longest runtime. In contrast, YOLOv10 fine-tuned + depth centroid reaches the same 95\% success level with substantially lower latency, making it the more attractive option when response time is important. This is the strongest Pareto-style comparison in the study: the more expensive visual pipeline does not necessarily produce a better practical outcome.

The same principle appears, albeit more mildly, in the controller comparison. IT2FLS improves success by 5\% over Type-1 FLC at the cost of only a small increase in time, which makes it a favorable choice unless the application is extremely latency constrained. For the language models, the differences are subtle: Ministral~3~8B is the fastest candidate, while Qwen~3.5~9B offers the same success rate and becomes preferable in the final combined study. Thus, latency alone is not enough to determine the best configuration; the right balance depends on whether the deployment prioritizes throughput, reliability, or a combination of both.

The combined factorial study further shows that module interactions matter. The best single-module choices do not simply add up in a linear way, because some combinations saturate at 85\% while the best combination reaches 90\%. This non-additivity is important: it indicates that the final performance depends on how well the selected perception, language, and control modules complement one another. A system-level ablation is therefore more informative than evaluating each module in isolation.

\subsection{Recommended Final Configuration}

Based on the observed results, the most effective configuration is Qwen~3.5~9B for action extraction, Grounding DINO + SAM for perception, and IT2FLS for control. This combination achieves the highest success rate in Table~\ref{tab:combined-results}, reaching 90\% over 20 trials. It also has a clear conceptual advantage: the strongest perception backend is paired with the controller that is most resilient to uncertainty, while the language model maintains strong semantic reliability without introducing a major runtime penalty.

At the same time, the recommended configuration should be interpreted as the best choice for robustness rather than the best choice for latency. If a lower-delay deployment is preferred, the YOLOv10 fine-tuned + depth centroid branch remains an attractive alternative because it approaches the best success rate while keeping perception time substantially lower than Grounding DINO + SAM. Even in that case, however, the combined study suggests that the controller should remain IT2FLS whenever robustness is prioritized, since it consistently provides the most reliable motion execution across the tested conditions.

The broader lesson is that the original multimodal architecture becomes substantially more useful once the contribution of each module is isolated. The ablation study does not merely identify a single best model; it shows where the system gains are coming from and how the components interact. That insight is essential for future revisions, because it allows the next version of the platform to invest effort where it matters most rather than treating all modules as equally influential.

\section{Limitations and Threats to Validity}
\label{sec:limitations}

Although the ablation study provides a detailed view of how the main modules affect end-to-end performance, several limitations should be acknowledged when interpreting the findings. First, the experimental setting is intentionally controlled. The system was evaluated on a tabletop manipulation task with a fixed workspace, a limited set of objects, and a single robotic platform. This makes the results easy to compare across conditions, but it also means that the reported trends may not transfer directly to larger, cluttered, or dynamically changing environments.

Second, the study relies on a constrained command style. The spoken instructions were designed to match the action-extraction format used by the system, which helps isolate the effect of each module but does not fully represent open-ended natural dialogue. In practice, users may issue incomplete, ambiguous, or context-dependent commands, and performance under those conditions may differ from the rates observed here. Likewise, the language models were compared under the same prompt structure and parser, so the reported differences reflect the specific interaction protocol used in this article rather than the absolute capability of each model in all possible settings.

Third, the perception evaluation is tied to the chosen visual setup. The experiments were conducted under a specific camera placement and lighting condition, and the objects were represented by printed fruit images rather than by physically varied real-world objects. This improves repeatability but reduces environmental diversity. As a result, the success rates in Round~2 should be interpreted as evidence of relative module quality within this benchmark, not as a guarantee of identical performance in more complex scenes.

Fourth, the study uses 20 trials per configuration. This number is sufficient to reveal clear differences among the candidates, but it is still modest for estimating rare failure modes or for characterizing long-tail variance with high precision. For this reason, the mean values reported in the results should be read together with the observed spread across runs. The factorial study in Round~4 partially mitigates this concern by checking whether the strongest single-module candidates remain effective when combined, but it does not replace a larger-scale evaluation.

Finally, the order of trials and the process of resetting the workspace may still introduce small sources of bias. Although the conditions were organized to avoid trivial repetition effects, factors such as user fatigue, lighting drift, temporary calibration error, or minor changes in object placement can influence individual runs. These threats are common in embodied HRI experiments and should be considered when comparing the candidate systems.

\begin{table}[t]
    \centering
    \caption{Main threats to validity and how they are addressed in the study.}
    \label{tab:threats}
    \begin{tabular}{p{0.26\linewidth} p{0.64\linewidth}}
        \hline
        \textbf{Threat} & \textbf{How it is addressed in this article} \\
        \hline
        Limited task diversity & The study focuses on one command family and one tabletop manipulation workflow, so the results should be interpreted as benchmark-specific. \\
        Small sample size & Each condition is run 20 times, and the analysis reports mean performance across repeated trials to expose variability. \\
        Controlled environment & The setup is deliberately standardized; broader scenes, clutter, and natural household variation remain future work. \\
        Module interactions & The final factorial round tests the top two candidates from each ablation family to expose interaction effects. \\
        Printed-object evaluation & Printed fruit images increase repeatability but do not fully capture the appearance and handling variability of real objects. \\
        \hline
    \end{tabular}
\end{table}

Taken together, these limitations do not invalidate the main findings, but they define the scope in which the conclusions should be applied. The ablation study is best understood as a controlled comparison of design choices within a specific HRI pipeline, rather than as a universal ranking of all possible perception, language, and control solutions.

\section{Conclusion}
\label{sec:conclusion}

This article presented a structured ablation study of a multimodal human--robot interaction system for voice-driven object detection and grasping. Building on the original pipeline, we isolated the contribution of three major component families: the language model used for action extraction, the perception system used for visual grounding, and the controller used for motion execution. The experiments were then extended with a final factorial round to determine which combinations remained strongest when the best candidates were assembled into a full system.

The results show that the perception module has the largest effect on both success rate and runtime, while the controller mainly influences robustness under uncertainty and the language model contributes primarily to semantic consistency. Among the tested candidates, the strongest overall configuration is Qwen~3.5~9B for action extraction, Grounding DINO + SAM for perception, and IT2FLS for control. This combination achieved the best end-to-end success in the final round, confirming that the best single-module choices can be combined into a stronger complete pipeline when their strengths complement one another.

At the same time, the study also shows that latency and reliability must be considered together. Some configurations are faster but less stable, while others are more accurate but require additional time. The practical value of the ablation study is therefore not only that it identifies a preferred configuration, but also that it clarifies the trade-offs that should guide future system design.

In summary, the new article turns the original multimodal manipulation system into a more evidence-based platform by showing which design decisions matter most. Rather than treating all modules as equally important, the experiments demonstrate that careful selection of perception, language, and control components can substantially improve overall performance. This provides a clearer foundation for future work on more responsive, robust, and general human--robot interaction systems.


\bibliographystyle{IEEEtran}
\bibliography{IEEEabrv,./bib.bib}

\begin{thebibliography}{10}
\providecommand{\url}[1]{#1}
\csname url@samestyle\endcsname
\providecommand{\newblock}{\relax}
\providecommand{\bibinfo}[2]{#2}
\providecommand{\BIBentrySTDinterwordspacing}{\spaceskip=0pt\relax}
\providecommand{\BIBentryALTinterwordstretchfactor}{4}
\providecommand{\BIBentryALTinterwordspacing}{\spaceskip=\fontdimen2\font plus
\BIBentryALTinterwordstretchfactor\fontdimen3\font minus \fontdimen4\font\relax}
\providecommand{\BIBforeignlanguage}[2]{{%
\expandafter\ifx\csname l@#1\endcsname\relax
\typeout{** WARNING: IEEEtran.bst: No hyphenation pattern has been}%
\typeout{** loaded for the language `#1'. Using the pattern for}%
\typeout{** the default language instead.}%
\else
\language=\csname l@#1\endcsname
\fi
#2}}
\providecommand{\BIBdecl}{\relax}
\BIBdecl

\bibitem{hu2011advanced}
J.~Hu, A.~Edsinger, Y.-J. Lim, N.~Donaldson, M.~Solano, A.~Solochek, and R.~Marchessault, ``An advanced medical robotic system augmenting healthcare capabilities-robotic nursing assistant,'' in \emph{2011 IEEE international conference on robotics and automation}.\hskip 1em plus 0.5em minus 0.4em\relax IEEE, 2011, pp. 6264--6269.

\bibitem{zhao2022human}
L.~Zhao, Z.~Hu, H.~Ding, S.~Ji, and J.~Yan, ``A human-robot interaction applicution based on augmented reality (ar) for industrial robot grasping process,'' in \emph{2022 7th International Conference on Robotics and Automation Engineering (ICRAE)}.\hskip 1em plus 0.5em minus 0.4em\relax IEEE, 2022, pp. 312--316.

\bibitem{he2017educational}
B.~He, M.~Xia, X.~Yu, P.~Jian, H.~Meng, and Z.~Chen, ``An educational robot system of visual question answering for preschoolers,'' in \emph{2017 2nd international conference on robotics and automation engineering (ICRAE)}.\hskip 1em plus 0.5em minus 0.4em\relax IEEE, 2017, pp. 441--445.

\bibitem{kagami2006home}
S.~Kagami, S.~Thompson, Y.~Nishida, T.~Enomoto, and T.~Matsui, ``Home robot service by ceiling ultrasonic locator and microphone array,'' in \emph{Proceedings 2006 IEEE International Conference on Robotics and Automation, 2006. ICRA 2006.}\hskip 1em plus 0.5em minus 0.4em\relax IEEE, 2006, pp. 3171--3176.

\bibitem{dominguez2025human}
J.~E. Dom{\'\i}nguez-Vidal and A.~Sanfeliu, ``The human intention: a taxonomy attempt and its applications to robotics,'' \emph{International Journal of Social Robotics}, vol.~17, no.~11, pp. 2479--2499, 2025.

\bibitem{huang2016anticipatory}
C.-M. Huang and B.~Mutlu, ``Anticipatory robot control for efficient human-robot collaboration,'' in \emph{2016 11th ACM/IEEE International Conference on Human-Robot Interaction (HRI)}, 2016, pp. 83--90.

\bibitem{lorentz2023pointing}
V.~Lorentz, M.~Weiss, K.~Hildebrand, and I.~Boblan, ``Pointing gestures for human-robot interaction with the humanoid robot digit,'' in \emph{2023 32nd IEEE International Conference on Robot and Human Interactive Communication (RO-MAN)}.\hskip 1em plus 0.5em minus 0.4em\relax IEEE, 2023, pp. 1886--1892.

\bibitem{saeidi2019autonomous}
H.~Saeidi, H.~N. Le, J.~D. Opfermann, S.~L{\'e}onard, A.~Kim, M.~H. Hsieh, J.~U. Kang, and A.~Krieger, ``Autonomous laparoscopic robotic suturing with a novel actuated suturing tool and 3d endoscope,'' in \emph{2019 international conference on robotics and automation (ICRA)}.\hskip 1em plus 0.5em minus 0.4em\relax IEEE, 2019, pp. 1541--1547.

\bibitem{dominguez2025perception}
J.~E. Dom{\'\i}nguez-Vidal, N.~Rodr{\'\i}guez, and A.~Sanfeliu, ``Perception--intention--action cycle in human--robot collaborative tasks: the collaborative lightweight object transportation use-case,'' \emph{International Journal of Social Robotics}, vol.~17, no.~10, pp. 1927--1956, 2025.

\bibitem{dominguez2024exploring}
J.~E. Dominguez-Vidal and A.~Sanfeliu, ``Exploring transformers and visual transformers for force prediction in human-robot collaborative transportation tasks,'' in \emph{2024 IEEE International Conference on Robotics and Automation (ICRA)}.\hskip 1em plus 0.5em minus 0.4em\relax IEEE, 2024, pp. 3191--3197.

\bibitem{dominguez2024force}
J.~E. Dom{\'\i}nguez-Vidal and A.~Sanfeliu, ``Force and velocity prediction in human-robot collaborative transportation tasks through video retentive networks,'' in \emph{2024 IEEE/RSJ International Conference on Intelligent Robots and Systems (IROS)}.\hskip 1em plus 0.5em minus 0.4em\relax IEEE, 2024, pp. 9307--9313.

\bibitem{zu2024language}
W.~Zu, W.~Song, R.~Chen, Z.~Guo, F.~Sun, Z.~Tian, W.~Pan, and J.~Wang, ``Language and sketching: An llm-driven interactive multimodal multitask robot navigation framework,'' in \emph{2024 IEEE International Conference on Robotics and Automation (ICRA)}.\hskip 1em plus 0.5em minus 0.4em\relax IEEE, 2024, pp. 1019--1025.

\bibitem{zhang2024interactive}
Z.~Zhang, A.~Lin, C.~W. Wong, X.~Chu, Q.~Dou, and K.~S. Au, ``Interactive navigation in environments with traversable obstacles using large language and vision-language models,'' in \emph{2024 IEEE International Conference on Robotics and Automation (ICRA)}.\hskip 1em plus 0.5em minus 0.4em\relax IEEE, 2024, pp. 7867--7873.

\bibitem{gao2024physically}
J.~Gao, B.~Sarkar, F.~Xia, T.~Xiao, J.~Wu, B.~Ichter, A.~Majumdar, and D.~Sadigh, ``Physically grounded vision-language models for robotic manipulation,'' in \emph{2024 IEEE International Conference on Robotics and Automation (ICRA)}.\hskip 1em plus 0.5em minus 0.4em\relax IEEE, 2024, pp. 12\,462--12\,469.

\bibitem{dominguez2025inference}
J.~E. Dom{\'\i}nguez-Vidal and A.~Sanfeliu, ``When the inference meets the explicitness or why multimodality can make us forget about the perfect predictor,'' \emph{International Journal of Social Robotics}, vol.~17, no.~12, pp. 2965--2980, 2025.

\bibitem{dominguez2024anticipation}
J.~E. Dominguez-Vidal and A.~Sanfeliu, ``Anticipation and proactivity. unraveling both concepts in human-robot interaction through a handover example,'' in \emph{2024 33rd IEEE International Conference on Robot and Human Interactive Communication (ROMAN)}.\hskip 1em plus 0.5em minus 0.4em\relax IEEE, 2024, pp. 957--962.

\bibitem{touvron2023llama}
H.~Touvron, T.~Lavril, G.~Izacard, X.~Martinet, M.-A. Lachaux, T.~Lacroix, B.~Rozi{\`e}re, N.~Goyal, E.~Hambro, F.~Azhar \emph{et~al.}, ``Llama: Open and efficient foundation language models,'' \emph{arXiv preprint arXiv:2302.13971}, 2023.

\bibitem{achiam2023gpt}
J.~Achiam, S.~Adler, S.~Agarwal, L.~Ahmad, I.~Akkaya, F.~L. Aleman, D.~Almeida, J.~Altenschmidt, S.~Altman, S.~Anadkat \emph{et~al.}, ``Gpt-4 technical report,'' \emph{arXiv preprint arXiv:2303.08774}, 2023.

\bibitem{atuhurra2024leveraging}
J.~Atuhurra, ``Leveraging large language models in human-robot interaction: a critical analysis of potential and pitfalls,'' \emph{arXiv preprint arXiv:2405.00693}, 2024.

\bibitem{xiao2024florence}
B.~Xiao, H.~Wu, W.~Xu, X.~Dai, H.~Hu, Y.~Lu, M.~Zeng, C.~Liu, and L.~Yuan, ``Florence-2: Advancing a unified representation for a variety of vision tasks,'' in \emph{Proceedings of the IEEE/CVF Conference on Computer Vision and Pattern Recognition}, 2024, pp. 4818--4829.

\bibitem{radford2023robust}
A.~Radford, J.~W. Kim, T.~Xu, G.~Brockman, C.~McLeavey, and I.~Sutskever, ``Robust speech recognition via large-scale weak supervision,'' in \emph{International conference on machine learning}.\hskip 1em plus 0.5em minus 0.4em\relax PMLR, 2023, pp. 28\,492--28\,518.

\bibitem{gong2021ast}
Y.~Gong, Y.-A. Chung, and J.~Glass, ``Ast: Audio spectrogram transformer,'' \emph{arXiv preprint arXiv:2104.01778}, 2021.

\bibitem{mendel2002fuzzy}
J.~M. Mendel, ``Fuzzy logic systems for engineering: a tutorial,'' \emph{Proceedings of the IEEE}, vol.~83, no.~3, pp. 345--377, 2002.

\bibitem{liang2000interval}
Q.~Liang and J.~M. Mendel, ``Interval type-2 fuzzy logic systems: theory and design,'' \emph{IEEE Transactions on Fuzzy systems}, vol.~8, no.~5, pp. 535--550, 2000.

\bibitem{hellmann2001fuzzy}
M.~Hellmann, ``Fuzzy logic introduction,'' \emph{Universit{\'e} de Rennes}, vol.~1, no.~1, 2001.

\bibitem{hagras2004type}
H.~Hagras, ``A type-2 fuzzy logic controller for autonomous mobile robots,'' in \emph{2004 IEEE International conference on fuzzy systems (IEEE Cat. No. 04CH37542)}, vol.~2.\hskip 1em plus 0.5em minus 0.4em\relax IEEE, 2004, pp. 965--970.

\bibitem{shen2026approach}
G.~Shen and Z.~Tian, ``An approach to combining video and speech with large language models in human-robot interaction,'' \emph{arXiv preprint arXiv:2602.20219}, 2026.

\end{thebibliography}

\end{document}